\documentclass[]{spie}  

\usepackage{amsmath,amsfonts,amssymb}
\usepackage{graphicx}
\usepackage[colorlinks=true, allcolors=blue]{hyperref}
\usepackage{hyperref}       
\usepackage{url}            
\usepackage{booktabs}       
\usepackage{amsfonts}
\usepackage{nicefrac}       
\usepackage{microtype}     
\usepackage{xcolor}   
\usepackage{graphicx}
\usepackage{footnote}
\usepackage{adjustbox}
\usepackage{pifont} 
\usepackage{subfigure}
\usepackage{xcolor} 
\usepackage{pifont} 

\title{Towards Synergistic Deep Learning Models for Volumetric Cirrhotic Liver Segmentation in MRIs}

\author{Vandan Gorade}
\author{Onkar Susladkar}
\author{Gorkem Durak}
\author{Elif Keles}
\author{Ertugrul Aktas}
\author{Timurhan Cebeci}
\author{Alpay Medetalibeyoglu}
\author{Daniela Ladner}
\author{Debesh Jha}
\author{Ulas Bagci}

\affil{Northwestern University, Chicago, IL, USA}

\begin{document} 
\maketitle

\begin{abstract}
Liver cirrhosis, a leading cause of global mortality, requires precise segmentation of ROIs for effective disease monitoring and treatment planning. Existing segmentation models often fail to capture complex feature interactions and generalize across diverse datasets. To address these limitations, we propose a novel synergistic theory that leverages complementary latent spaces for enhanced feature interaction modeling. Our proposed architecture, \textit{nnSynergyNet3D}, integrates continuous and discrete latent spaces for 3D volumes and features auto-configured training. This approach captures both fine-grained and coarse features, enabling effective modeling of intricate feature interactions. We empirically validated \textit{nnSynergyNet3D} on a private dataset of 628 high-resolution T1 abdominal MRI scans from 339 patients. Our model outperformed the baseline \textit{nnUNet3D} by approximately 2\%. Additionally, zero-shot testing on healthy liver CT scans from the public LiTS dataset demonstrated superior cross-modal generalization capabilities. These results highlight the potential of synergistic latent space models to improve segmentation accuracy and robustness, thereby enhancing clinical workflows by ensuring consistency across CT and MRI modalities.

\end{abstract}

\section{Introduction}

\textbf{{Clinical Motivation}.} Liver cirrhosis, the final stage of chronic liver disease (CLD), represents a significant global health issue. In 2019, it ranked as the 11th leading cause of death, contributing to 2.4\% of deaths worldwide~\cite{GINES20211359,naturereview}. While viral hepatitis remains the primary cause of end-stage liver disease, metabolic dysfunction-associated steatotic liver disease (MASLD) is projected to surpass it as the leading cause, driven by the rising prevalence of obesity and metabolic syndrome globally~\cite{naturereview}. Additionally, conditions such as alcoholic liver disease, autoimmune hepatitis, and genetic disorders also play substantial roles~\cite{naturereview}. Cirrhosis involves the development of bridging fibrosis and regenerative nodules, which compromise liver function and can lead to liver failure~\cite{pellicoro}. Accurate segmentation of cirrhotic livers in radiology images is essential for monitoring the disease's progression, assessing severity, and evaluating treatment responses. Precise segmentation helps determine the extent and location of liver damage, which is vital for planning treatments like liver transplantation and other targeted therapies.

\noindent\textbf{{Theoretical Shortcomings}.} Existing segmentation models often face significant theoretical shortcomings, primarily due to their limited ability to capture complex feature interactions and their fixed hypothesis spaces\cite{kawaguchi2017generalization}. Traditional models\cite{chen2021transunet,isensee2021nnu,hatamizadeh2021swin} treat features in isolation or use simplistic integration methods, failing to exploit intricate, non-linear dependencies between different feature sets. This results in suboptimal representation of complex patterns, such as subtle tissue variations in MRI scans. Additionally, the fixed nature of these models' hypothesis spaces restricts their generalization capacity\cite{kawaguchi2017generalization} to new, diverse datasets, limiting their robustness across varying imaging conditions. Thus, these limitations can impede the accuracy and adaptability of conventional segmentation approaches in challenging scenarios like cirrhotic liver segmentation.

\noindent\textbf{Synergistic Theory and Practical Architecture Design.}To address these limitations, we propose a novel synergistic theory that enhances feature interaction modeling by incorporating features from diverse latent spaces that complement each other. This theory posits that leveraging synergistic representations\cite{gorade2024synergynet,gorade2024harmonized} where different types of latent features interact and complement each other can more effectively capture complex, non-linear relationships within the data. Based on this synergistic theory, we introduce the \textit{nnSynergyNet3D} architecture, which integrates both continuous and discrete latent spaces for 3D volumes and enjoys auto-configured training. The continuous latent space captures fine-grained details, while the discrete latent space addresses broader, coarse features. This dual approach allows nnSynergyNet3D to model intricate feature interactions more effectively. Further, auto-configuration\cite{isensee2021nnu} allows the model to select the best settings and parameters automatically. This combination of automatic configuration and synergistic information allows for more effective modeling of complex patterns, enhancing the accuracy and robustness of segmentation, particularly in challenging scenarios like cirrhotic liver segmentation.

\noindent\textbf{Empirical Analysis}
We quantitatively and qualitatively validate the effectiveness of nnSynergyNet3D on the task of cirrhotic liver segmentation, where capturing both discrete and continuous features is crucial. Extensive experiments on our private dataset, consisting of 628 high-resolution T1 abdominal MRI volumetric scans from 339 patients with corresponding physician-annotated segmentation masks, demonstrate that nnSynergyNet3D consistently outperforms the strong baseline method nnUNet3D by approximately 2\%. Additionally, we assess the cross-modal generalization capability of our method by performing zero-shot testing on healthy liver CT scans from the publicly available LiTS\cite{heller2019kits19} dataset. Our results reveal that nnSynergyNet3D shows superior generalization capabilities compared to existing methods, suggesting that leveraging synergistic latent space models could significantly benefit clinical workflows by improving segmentation consistency across both CT and MRI modalities, ultimately easing the workload for clinicians. Overall, our theoretical guarantees backed by empirical results highlight the importance of synergistic representations in enhancing the accuracy and robustness of segmentation models.


\section{Methodology}

\subsection{Problem Statement}
Given a set of MRI scans \( \mathcal{X} \subseteq \mathbb{R}^d \), our goal is to segment cirrhotic liver tissues accurately. Let \( \mathbf{x} \in \mathcal{X} \) be an input scan and \( \mathbf{y} \in \mathcal{Y} \) be the corresponding segmentation map. Traditional segmentation methods often struggle with the complex variations present in MRI images. We address this by proposing a deep learning model that leverages synergistic feature spaces.

\subsection{Synergistic Latent Space Theory}

\begin{figure}  [!t]
    \centering
    \includegraphics[width=0.8\textwidth]{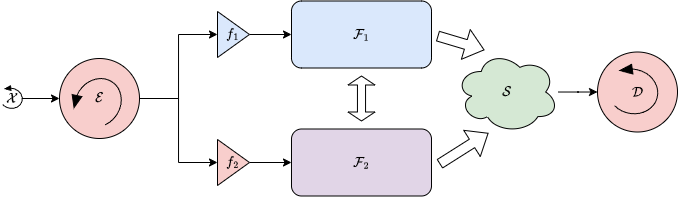}
    \caption{\textbf{Visual Representation of Proposed Synergistic Latent Space Methodology}: In the diagram, \( f_1 \) and \( f_2 \) represent transformation functions that map one feature space to another. The circular arrows indicate auto-configuration processes. \( F_1 \) and \( F_2 \) denote complementary feature spaces, while \( S \) represents the synergistic latent space that integrates and enhances the information from \( F_1 \) and \( F_2 \).}

    \label{qualitative-t1-t2}
\end{figure}

We introduce two feature sets, \( \mathbf{F}_1 \) and \( \mathbf{F}_2 \), derived from the MRI scans. These feature sets are combined to form a synergistic space \( \mathcal{S} = \mathbf{F}_1 \times \mathbf{F}_2 \). For the segmentation function \( f: \mathcal{S} \to \mathbb{R} \), we aim to prove that our model can approximate \( f \) accurately and generalize well to unseen data. To establish this, we first need to demonstrate that a neural network can effectively approximate any function within this synergistic feature space.

\noindent\textbf{Theorem 1:} Synergistic Latent Approximation Theorem. Let \( f: S \to \mathbb{R} \) be a continuous function on a compact set \( S \subset \mathbb{R}^{d_1 + d_2} \). If \( \mathcal{N} \) is an algebra of functions on \( \mathbb{R}^{d_1 + d_2} \) that separates points and contains constant functions, then for any \( \epsilon > 0 \), there exists \( g \in \mathcal{N} \) such that:
\[
\| f - g \|_\infty < \epsilon.
\]
\noindent\textbf{Proof:} By the Stone-Weierstrass Theorem~\cite{de1959stone,cotter1990stone}, the algebra \( \mathcal{N} \), which includes neural network functions capable of modeling complex synergistic features, is dense in \( C(S) \). Therefore, for any continuous function \( f \) and any \( \epsilon > 0 \), there exists a function \( g \in \mathcal{N} \) such that:
\[
\| f - g \|_\infty < \epsilon.
\]
The capacity of neural networks to model feature interactions ensures that these approximations can capture intricate relationships within the data. For a detailed proof, please refer to the supplementary materials.

\noindent\textbf{Theorem 2:} Generalization Bound. Let \( \mathcal{H} \) be the hypothesis space of a deep learning model trained on \( \mathcal{S} \). The generalization error \( |R(f) - R_{emp}(f)| \) can be bounded as:
\[
|R(f) - R_{emp}(f)| \leq \mathcal{O} \left( \sqrt{\frac{VC(\mathcal{H}) \log(N)}{N}} \right)
\]

\noindent\textbf{Proof:} To bound the generalization error \( |R(f) - R_{emp}(f)| \), we leverage Hoeffding's inequality\cite{bentkus2004hoeffding} and Rademacher complexity\cite{bartlett2002rademacher,kawaguchi2017generalization}. For a model utilizing the synergistic feature space \( \mathcal{S} = \mathbf{F}_1 \times \mathbf{F}_2 \), the generalization error is bounded by:
\[
|R(f) - R_{emp}(f)| \leq \sqrt{\frac{VC(\mathcal{H}_{\mathbf{F}_1}) + VC(\mathcal{H}_{\mathbf{F}_2}) \log(N)}{N}},
\]
where \( VC(\mathcal{H}_{\mathbf{F}_1}) \) and \( VC(\mathcal{H}_{\mathbf{F}_2}) \) are the VC dimensions\cite{vapnik1991necessary} of the feature spaces \( \mathbf{F}_1 \) and \( \mathbf{F}_2 \), respectively. This bound indicates that the enriched hypothesis space of the synergistic model improves its capacity to fit complex functions while ensuring effective generalization. For a detailed proof, please refer to the supplementary materials.

\subsection{Practical Architecture Design}
Based on the synergistic latent space theory, we propose a deep learning architecture that leverages interactions between complementary spaces, named \textit{nnSynergyNet3D}. Unlike the traditional 2D SynergyNet\cite{gorade2024synergynet}, our design employs a 3D cascade U-Net. Initially, a 3D U-Net processes low-resolution images, followed by a second high-resolution 3D U-Net that refines the predictions of the former. At the bottleneck, we use \( \mathcal{F}_1 \) and \( \mathcal{F}_2 \) as continuous and discrete features, respectively, to capture fine and coarse information in cirrhotic liver MRI scans. Discrete features are learned using volumetric vector quantization, which dynamically discretizes continuous 3D features. Finally, \( \mathcal{F}_1 \) and \( \mathcal{F}_2 \) are combined using volumetric multi-head cross-attention to capture synergistic information. The SynergyNet3D bottleneck integrates three-dimensional continuous and discrete latent spaces to capture intricate, non-linear feature interactions effectively. Inspired by nnU-Net~\cite{isensee2021nnu}, nnSynergyNet3D employs rule-based parameters to adapt network topology, patch size, and batch size based on the dataset fingerprint and GPU memory constraints using hard-coded heuristic rules. Empirical parameters, determined through trial and error, are used to select the best U-Net configuration (e.g. 3D full resolution, 3D low resolution, 3D cascade) and optimize the postprocessing strategy. This combination of automatic configuration and synergistic information allows for more effective modeling of complex patterns, enhancing the accuracy and robustness of segmentation, particularly in challenging scenarios like cirrhotic liver segmentation.

\section{Experimental Setup}

We carefully selected several baseline methods such as nnUNet3D~\cite{isensee2021nnu}, nnFormer3D~\cite{zhou2023nnformer}, and TransUNet3D~\cite{chen2021transunet} etc.

\noindent\textbf{Datasets.} 
We leveraged our private data cirrhotic liver MRI data and LiTS\cite{heller2019kits19} datasets to demonstrate the effectiveness of our proposed method. Our private  data includes 628 high-resolution abdominal MRI scans, consisting of 310 T1-weighted (T1W) and 318 T2-weighted (T2W) volumetric scans from 339 patients. It encompasses both contrast-enhanced and non-enhanced MRI scans, along with segmentation masks annotated by physicians. This is a comprehensive single-center dataset that is multivendor, multiplanar, and multiphase. The dataset was split with an 80:10:10 ratio, resulting in 248 cases for training, 31 cases for validation, and 31 cases for testing for T1W. Similarly, for T2W, we allocated 256 cases for training, and 31 cases each for validation and testing. While the T2W split was not exactly 80:10:10, we maintained the distribution as close to the target ratio as possible. This careful balancing accounts for the domain shift introduced by different device vendors, resulting in variable scan characteristics across splits. 

\noindent\textbf{Metrics.} To assess the liver segmentation performance, we utilized several metrics, including the mean Dice Similarity Coefficient (mDSC), mean Intersection over Union (mIoU), recall, precision, Hausdorff Distance (HD95), and Average Symmetric Surface Distance (ASSD). These metrics provide a comprehensive evaluation of segmentation accuracy, offering insights into both the overlap and boundary precision of the segmented regions

\noindent\textbf{Implementation Details.} We trained our models using PyTorch 2.2.2 with CUDA 11.2. The learning rate was initially set to 0.0001 and was gradually decreased using the Cosine Annealing Scheduler. We use the BCE-Dice loss with the AdamW optimizer with a batch size of 4. The models were trained for 500 epochs with an early stopping patience of 50 to prevent overfitting. The learning rate decay was set at 0.001 after every 10 epochs. To accelerate training, we leveraged two Nvidia A6000 GPUs, each with 48GB of memory, and utilized PyTorch's Distributed Data Parallel to distribute a batch of 4 to each GPU. We resized every volume to uniform spatial dimensions of $256\times256\times80$ for generalizability.

\section{Empirical Analysis}

\begin{table}[!t]
\caption{Comparative benchmark of SOTA 3D segmentation networks across various metrics on T1w Liver Cirrhosis MRI dataset. Bold shows the best performance while red is the second-best.}
\centering
\footnotesize
\begin{adjustbox}{max width=\textwidth}
\begin{tabular}{lccccccc}
\toprule
\textbf{Method} & \textbf{mIoU} & \textbf{Dice} & \textbf{HD95} & \textbf{Precision} & \textbf{Recall} & \textbf{ASSD}   \\ 
\midrule

\texttt{nnUNet3D}~[\cite{zhou2023nnformer}]  & 82.22 & 85.72 & 26.78 & 86.67 & 85.98 & 4.38   \\ 

\texttt{nnFormer3D}~[\cite{zhou2023nnformer}]  & 83.03 & 86.09 & \textcolor{red}{25.18} & \textcolor{red}{87.11} & 85.72 & \textcolor{red}{4.01}   \\  
\texttt{TransUNet3D}~[\cite{chen2021transunet}] & 79.19 & 80.92 & 31.09 & 80.01 & 79.91 & 5.92  \\ 
\texttt{SwinUNeTr}~[\cite{hatamizadeh2021swin}] & 81.02 & 82.01 & 30.66 & 81.32 & 80.97 & 5.01   \\ 
\texttt{TransBTS}~[\cite{wenxuan2021transbts}] & 63.42 & 76.11 & 36.92 & 74.84 & 84.01 & 7.39   \\ 
\hline
\texttt{nnSynergyNet3D}~[\cite{gorade2024synergynet}] & \textbf{84.51} & \textbf{87.89} & \textbf{21.04} & \textbf{88.72} & 87.76 & \textcolor{red}{4.01}  \\

 \texttt{w/o Autoconf.}~[\cite{gorade2024synergynet}] & 76.11 & 78.77 & 27.55 & 85.12 & 86.72 & 5.34  \\ 

\bottomrule
\end{tabular}
\end{adjustbox}
\label{tab:T1segmentation_methods}
\end{table}

\noindent\textbf{Quantitative analysis on T1W modality.} Table~\ref{tab:T1segmentation_methods} provides a comparison between our proposed \textit{nnSynergyNet3D} and five state-of-the-art (SOTA) segmentation networks on the T1W dataset. Our \textit{nnSynergyNet3D} outperforms all comparison methods across all metrics. Among the transformer-based methods, \textit{nnFormer3D} showed comparable performance in capturing cirrhotic liver tissue and its boundaries, highlighting the significance of long-range dependencies. However, models such as \textit{SwinUNeTr} and \textit{TransUNet3D} did not significantly surpass CNN-based models like \textit{nnUNet}, emphasizing the necessity of more advanced configurations. Moving to transformer-based models with auto-configuration, \textit{nnFormer3D} and our \textit{nnSynergyNet3D} demonstrated superior performance due to their ability to adapt to the liver's varying shapes and complex boundaries. The auto-configured continuous and discrete representation enabled these models to capture fine and coarse features more effectively. Our proposed \textit{nnSynergyNet3D} achieved the highest overall performance, with a mean Intersection over Union (mIoU) of 84.51, Dice Similarity Coefficient (DSC) of 87.89\%, Hausdorff Distance at 95th percentile (HD95) of 21.04 mm, and precision of 88.72\%. This superior performance can be attributed to the synergistic integration of continuous and discrete representations, which allowed the model to capture both fine details and long-range dependencies due to its Transformer-inspired design. This demonstrates the critical role of combining transformer-based architectures with auto-configuration for enhancing segmentation performance in complex tasks like cirrhotic liver segmentation.

\noindent\textbf{Quantitative analysis on T2w modality} Table~\ref{tab:T2segmentation_methods} presents an evaluation of state-of-the-art (SOTA) 3D segmentation networks on the T2W dataset scans. The results closely mirror those observed in the T1W segmentation results. Among the evaluated models, our \textit{nnSynergyNet3D} stands out with a superior Dice Similarity Coefficient (DSC) value of 86.51\%, the lowest Hausdorff Distance (HD) of 24.19 mm, and the lowest Average Surface Distance (ASD) value of 3.96 mm. These metrics underscore the effectiveness of our approach, particularly in handling the T2-weighted (T2W) MRI characteristics, which typically provide enhanced contrast between different soft tissues and are crucial for identifying cirrhotic liver tissues. \textit{nnFormer3D} and \textit{nnUNet3D} also demonstrated competitive performance, reinforcing the importance of transformer-based designs for capturing intricate details of cirrhotic liver tissues. T2W images often highlight fibrotic changes and fluid accumulation, which are critical for accurate cirrhotic liver segmentation. However, models such as \textit{SwinUNeTr}, \textit{TransBTS}, and \textit{TransUNet3D} did not significantly surpass the performance of CNN-based models like \textit{nnUNet} and \textit{SynergyVNet3D}. This emphasizes the critical role of auto-configured and hybrid CNN-Transformer-based models in achieving competitive segmentation results, particularly for the complex structures and varying intensities in T2W images.

\begin{table}[!ht]
    \caption{Comparative benchmark of SOTA 3D segmentation networks on T2W. Bold shows the best performance while red is the second-best.} 
    \centering
    \footnotesize
    \begin{adjustbox}{max width=\textwidth}
\begin{tabular}{lccccccc}
    \toprule
\textbf{Method}	&  \textbf{mIoU}	& \textbf{Dice}& \textbf{HD95} & \textbf{Precision} & \textbf{Recall}	& \textbf{ASDD}  &  \\
\midrule    

\texttt{nnUNet3D}~[\cite{isensee2021nnu}] 	&82.11	&84.76	&27.73	&\textcolor{red}{85.78}	&86.66	&4.76\\

\texttt{nnFormer3D}~[\cite{zhou2023nnformer}]	&\textbf{83.42}	&\textcolor{red}{86.47}	&\textcolor{red}{25.92}	&\textbf{87.67}	&\textbf{88.02}	&\textcolor{red}{4.04}\\



\texttt{TransUNet3D}~[\cite{chen2021transunet}] 	&77.89	&79.09	&34.11	&78.11	& 79.97	&6.69 \\


\texttt{SwinUNeTr}~[\cite{hatamizadeh2021swin}]	&79.89	&81.21	&32.78	&80.05	&81.10	&6.19\\

\texttt{TransBTS}~[\cite{wenxuan2021transbts}] &62.80  &74.88 &43.73 &  76.69   & 79.75  & 8.18   \\ 

\hline
\texttt{nnSynergyNet3D}~[\cite{gorade2024synergynet}]	&\textcolor{red}{83.01}	&\textbf{86.51}	&\textbf{24.19}	&85.66	&\textcolor{red}{87.01}	&\textbf{3.96} \\

\texttt{w/o Autoconf.}~[\cite{gorade2024synergynet}] & 75.17 & 77.56 &	28.19 &	83.78 &	85.42 &	5.79 \\

\bottomrule
\end{tabular}
\end{adjustbox}
\label{tab:T2segmentation_methods}
\end{table}

\begin{table}[t]
    \caption{Cross-domain analysis of the model trained on T1W and tested on LiTS. Bold shows the best performance while red is the second-best.} 
    \centering
    \footnotesize
    \begin{adjustbox}{max width=\textwidth}
    \begin{tabular}{lccccccc}
        \toprule
        \textbf{Method} & \textbf{mIoU} & \textbf{Dice} & \textbf{HD95} & \textbf{Precision} & \textbf{Recall} & \textbf{ASSD}  \\
        \midrule
        \texttt{nnUNet3D}~[\cite{isensee2021nnu}] & 80.02 & 81.19 & 29.78 & 80.98 & 83.21 & 6.52   \\
         \texttt{nnFormer3D}~[\cite{zhou2023nnformer}] & \textcolor{red}{80.41} & \textcolor{red}{82.19} & \textcolor{red}{29.19} & \textbf{82.76} & \textcolor{red}{85.33} & \textcolor{red}{5.91}   \\
         \hline
         \texttt{nnSynergyNet3D}~[\cite{gorade2024synergynet}] & \textbf{81.88} & \textbf{83.38} & \textbf{29.01} & \textcolor{red}{81.15} & \textbf{86.02} & \textbf{5.55}  \\
        \bottomrule
    \end{tabular}
    \end{adjustbox}
    \label{tab:crossdomain}
\end{table}

\noindent\textbf{Cross-domain analysis:} Table~\ref{tab:crossdomain} presents the results of segmentation networks pre-trained on T1W MRIs and evaluated on CT scans from the LiTS dataset. The findings demonstrate that \textit{nnSynergyNet3D} for training on the T1W dataset enhances the model's ability to generalize to liver CT scans. This suggests that leveraging synergistic latent space models could significantly benefit clinical workflows by improving the consistency of segmentation across both CT and MRI modalities, ultimately easing the workload for clinicians. 

\begin{figure}[!ht]
    \centering
    \includegraphics[width=0.95\textwidth]{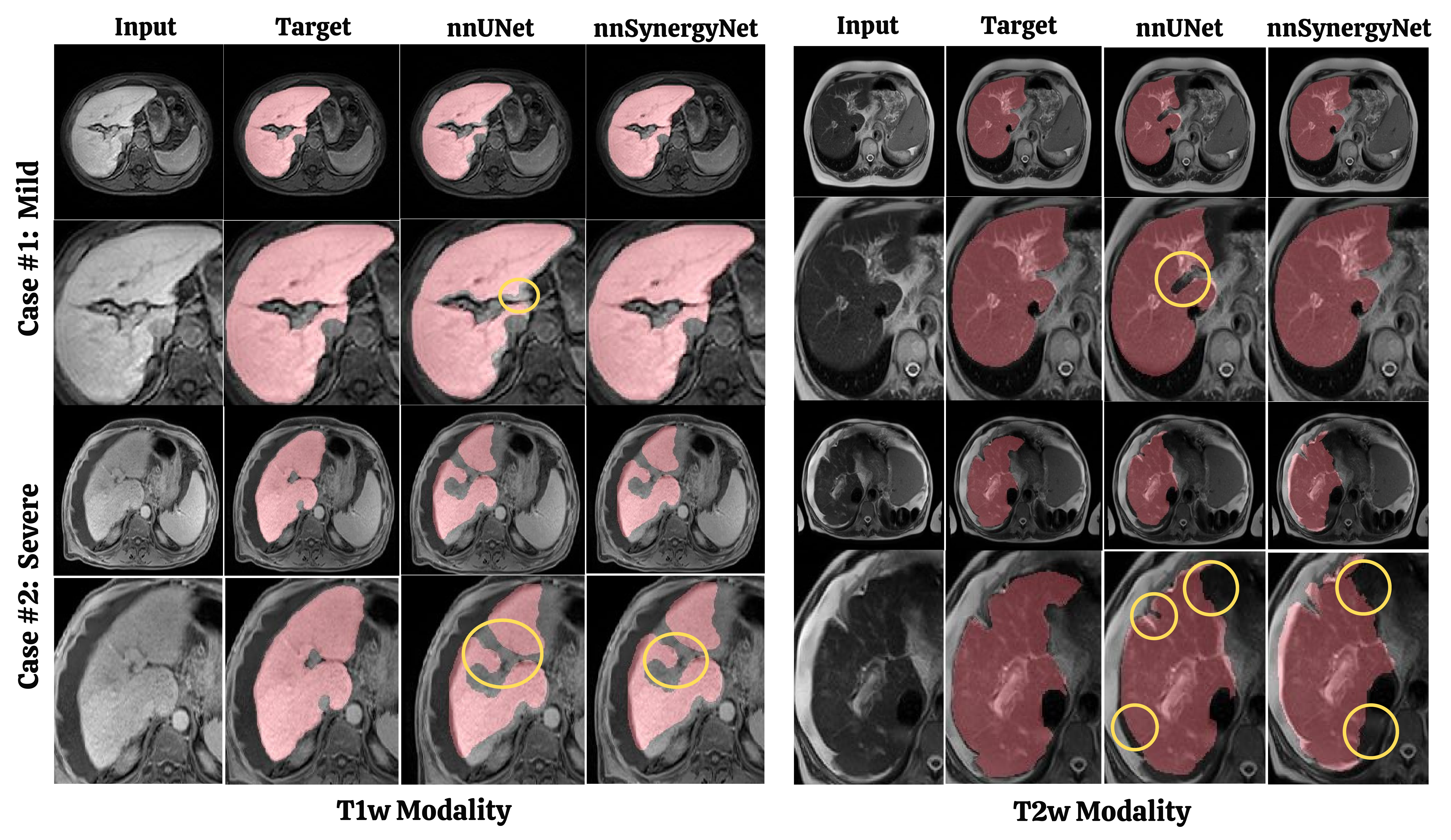}
    \caption{Qualitative comparison between nnsynergyNet3D and nnUNet on segmenting mild and severe cirrhosis from abdominal T1-weighted and  T2-weighted MRI scans. The yellow bounding circles show major mistakes made by the networks.}
    \label{qualitative-t1-t2}
\end{figure}

\noindent\textbf{Qualitative Analysis.}
Fig.~\ref{qualitative-t1-t2} displays T1W and T2W sample results, showing that current models segment cirrhotic livers well under mild conditions. However, the white boundaries indicate these models struggle with moderate-to-severe cases due to scarring. \textit{nnSynergyNet3D} consistently outperforms \textit{nnUNet, TransUNet} even in advanced cases, demonstrating its auto-configuration coupled with synergistic latent space design.

\section{Conclusion}

In this study, we introduced a novel synergistic theory and the \textit{nnSynergyNet3D} architecture to address the limitations of existing segmentation models in capturing complex feature interactions and generalizing across diverse datasets. By integrating continuous and discrete latent spaces, nnSynergyNet3D effectively models intricate feature interactions, enhancing segmentation accuracy and robustness. Our empirical validation on a comprehensive dataset of T1 abdominal MRI scans demonstrated that nnSynergyNet3D outperforms the state-of-the-art nnUNet3D by approximately 2\%. Furthermore, our model exhibited superior cross-modal generalization capabilities in zero-shot testing on healthy liver CT scans. These findings underscore the potential of synergistic latent space models to significantly benefit clinical workflows, particularly in challenging scenarios like cirrhotic liver segmentation, by improving segmentation consistency across both CT and MRI modalities. Future work will explore further optimization and extension of synergistic representations to other medical imaging tasks.

\acknowledgments 
This research is funded by NIH grants R01-CA246704, R01-CA240639, U01DK127384-02S1, and U01-CA268808.

\section{Appendix}

\subsection{Detailed Proofs}
\noindent\textbf{Theorem 1:} Synergistic Latent Approximation Theorem. Let \( f: S \to \mathbb{R} \) be a continuous function on a compact set \( S \subset \mathbb{R}^{d_1 + d_2} \). If \( \mathcal{N} \) is an algebra of functions on \( \mathbb{R}^{d_1 + d_2} \) that separates points and contains constant functions, then for any \( \epsilon > 0 \), there exists \( g \in \mathcal{N} \) such that:
\[
\| f - g \|_\infty < \epsilon.
\]
\noindent\textbf{Detailed Proof.} Following Stone-Weierstrass Theorem, which states that if \( A \) is an algebra of continuous functions on a compact set \( S \subset \mathbb{R}^d \) and \( A \) satisfies two conditions—point separation and inclusion of constant functions—then the closure of \( A \) in the supremum norm \( \| \cdot \|_{\infty} \) equals \( C(S) \), the space of all continuous functions on \( S \). Formally,
\[
\overline{A} = C(S),
\]
where \( \overline{A} \) denotes the closure of \( A \) in \( \| \cdot \|_{\infty} \).
To apply this theorem to our context, let \( N \) represent the algebra of functions computed by neural networks with continuous and non-constant activation functions. This algebra \( N \) separates points and contains constant functions, fulfilling the criteria stipulated by the Stone-Weierstrass Theorem. Consequently,
\[
\overline{N} = C(S).
\]
This implies that \( N \) is dense in \( C(S) \) with respect to the supremum norm \( \| \cdot \|_{\infty} \). Hence, for any continuous function \( f \) defined on \( S \) and for any \( \epsilon > 0 \), there exists a function \( g \in N \) such that:
\[
\| f - g \|_{\infty} < \epsilon.
\]
The significance of this result lies in the ability of neural networks to approximate any continuous function \( f \) on \( S \) with arbitrary precision. The neural network's algebra \( N \), which incorporates interactions between features, enhances the expressiveness of the function class, allowing it to capture complex dependencies between feature sets. This approximation capability is crucial in practical scenarios where modeling intricate relationships between features is required. Thus, the theorem demonstrates that any continuous function \( f \) defined on a compact set \( S \subset \mathbb{R}^{d_1 + d_2} \) can be approximated as closely as desired by a function \( g \) in \( N \), ensuring:
\[
\| f - g \|_{\infty} < \epsilon.
\]
\noindent\textbf{Theorem 2:} Generalization Bound. Let \( \mathcal{H} \) be the hypothesis space of a deep learning model trained on \( \mathcal{S} \). The generalization error \( |R(f) - R_{emp}(f)| \) can be bounded as:
\[
|R(f) - R_{emp}(f)| \leq \mathcal{O} \left( \sqrt{\frac{VC(\mathcal{H}) \log(N)}{N}} \right)
\]

\noindent\textbf{Detailed Proof.} To establish the generalization bound for our model, we start by analyzing the generalization error defined as the difference between the true risk $R(f)$ and the empirical risk $R_{\text{emp}}(f)$. The true risk $R(f)$ is given by:
\[
R(f) = \mathbb{E}_{z \sim D}[\ell(f(z), y)],
\]
where $\ell$ denotes the loss function and $D$ represents the underlying data distribution. The empirical risk $R_{\text{emp}}(f)$ is defined as:
\[
R_{\text{emp}}(f) = \frac{1}{N} \sum_{i=1}^{N} \ell(f(z_i), y_i),
\]
where $(z_i, y_i)$ are the training samples and $N$ is the number of samples. We use Hoeffding’s inequality to bound the deviation between $R(f)$ and $R_{\text{emp}}(f)$. Hoeffding's inequality states that for any $\epsilon > 0$:
\[
\Pr\left(\left|R(f) - R_{\text{emp}}(f)\right| \geq \epsilon \right) \leq 2 \exp\left(-\frac{2 N \epsilon^2}{\Delta^2}\right),
\]
where $\Delta$ is the range of the loss function $\ell$. This inequality implies that with high probability, the difference between the true and empirical risk is bounded by $\epsilon$, provided $N$ is sufficiently large. Next, we utilize the concept of Rademacher complexity to derive a more specific bound. The Rademacher complexity of the hypothesis space $H$, denoted as $R_N(H)$, measures the capacity of $H$ to fit random noise. It is known that:
\[
R_N(H) \leq \frac{VC(H) \log(N)}{N},
\]
where $VC(H)$ is the VC dimension of the hypothesis space $H$. This relation shows that the complexity of $H$ affects the generalization error, with a larger VC dimension implying a higher capacity to fit data. For our synergistic feature space $S$, which is defined as $S = F_1 \times F_2$, the hypothesis space $H$ formed from $S$ can be characterized by:
\[
VC(H) \geq VC(H_{F_1}) + VC(H_{F_2}),
\]
where $H_{F_1}$ and $H_{F_2}$ are the hypothesis spaces associated with the feature sets $F_1$ and $F_2$, respectively. This inequality reflects that the VC dimension of $H$ in the synergistic space $S$ is at least the sum of the VC dimensions of the individual feature spaces. This implies that the hypothesis space $H$ is richer and can represent more complex functions due to the interaction between $F_1$ and $F_2$. Combining Hoeffding’s inequality with the Rademacher complexity, we obtain:
\[
\left|R(f) - R_{\text{emp}}(f)\right| \leq R_N(H) + \frac{\log(1/\delta)}{2 N}.
\]
For large $N$, the term $\frac{\log(1/\delta)}{2 N}$ becomes negligible. Hence, the generalization error can be bounded primarily by:
\[
\left|R(f) - R_{\text{emp}}(f)\right| \leq R_N(H).
\]
Substituting the bound for Rademacher complexity, we get:
\[
\left|R(f) - R_{\text{emp}}(f)\right| \leq \frac{VC(H) \log(N)}{N}.
\]
Specifically, for the synergistic space $S$, we substitute $VC(H)$ with $VC(H_{F_1}) + VC(H_{F_2})$:
\[
\left|R(f) - R_{\text{emp}}(f)\right| \leq \frac{VC(H_{F_1}) + VC(H_{F_2}) \log(N)}{N}.
\]
This bound demonstrates that the increased VC dimension in the synergistic space $S$ enhances the hypothesis space's capacity to capture more complex functions. Consequently, this increased capacity improves the model’s ability to fit the training data, while the generalization bound ensures that this capacity translates into better performance on unseen data. Thus, the theoretical analysis confirms that using a synergistic feature space not only enhances the model's ability to represent complex functions but also ensures effective generalization, given sufficient training data.

\bibliography{report} 
\bibliographystyle{spiebib} 

\end{document}